\documentclass[preprint,12pt]{elsarticle}
\usepackage{amsmath}
\usepackage{booktabs}
\usepackage{amssymb}
\usepackage{multirow}
\usepackage{bm}
\usepackage[linesnumbered,ruled,vlined]{algorithm2e}
\usepackage{listings, xcolor}
\usepackage{natbib}

\newtheorem{definition}{Definition}

\journal{Neural Networks}

\begin{document}

\begin{frontmatter}

\title{Attributed Multi-order Graph Convolutional Network for Heterogeneous Graphs}

\author[label1,label2]{Zhaoliang~Chen}
\author[label1,label2]{Zhihao~Wu}
\author[label1,label2]{Luying~Zhong}
\author[label3,label4]{Claudia~Plant}
\author[label1,label2]{Shiping~Wang}
\author[label1,label2]{Wenzhong~Guo\corref{cor1}}
\ead{guowenzhong@fzu.edu.cn}
\cortext[cor1]{Corresponding author.}

\address[label1]{College of Computer and Data Science, Fuzhou University, Fuzhou 350116, China}
\address[label2]{Fujian Provincial Key Laboratory of Network Computing and Intelligent Information Processing, Fuzhou University, Fuzhou 350116, China}
\address[label3]{Faculty of Computer Science, University of Vienna, Vienna 1090, Austria}
\address[label4]{ds:UniVie, Vienna 1090, Austria}

\begin{abstract}
  Heterogeneous graph neural networks aim to discover discriminative node embeddings and relations from multi-relational networks.
  One challenge of heterogeneous graph learning is the design of learnable meta-paths, which significantly influences the quality of learned embeddings.
  Thus, in this paper, we propose an Attributed Multi-Order Graph Convolutional Network (AMOGCN), which automatically studies meta-paths containing multi-hop neighbors from an adaptive aggregation of multi-order adjacency matrices.
  The proposed model first builds different orders of adjacency matrices from manually designed node connections.
  After that, an intact multi-order adjacency matrix is attached from the automatic fusion of various orders of adjacency matrices.
  This process is supervised by the node semantic information, which is extracted from the node homophily evaluated by attributes.
  Eventually, we utilize a one-layer simplifying graph convolutional network with the learned multi-order adjacency matrix, which is equivalent to the cross-hop node information propagation with multi-layer graph neural networks.
  Substantial experiments reveal that AMOGCN gains superior semi-supervised classification performance compared with state-of-the-art competitors.
\end{abstract}


\end{frontmatter}

\section{Introduction}
Graphs are universally available in practical applications, e.g., links between websites, paper citations, clicks and purchases of users.
As a powerful tool to process graph-structured data, graph neural networks have been extensively investigated in recent years \cite{PanCH23,XuXGHG21,YanhuiHybrid2022,XiaWGYG23,XiaWYGHG22}.
For instance, Graph Convolutional Network (GCN) \cite{KipfW17} and its variants have been widely leveraged in a large number of models owing to its excellent ability to capture discriminative multi-hop node embeddings from neighbors \cite{zhang2021shne,LiXCZL21,XiaWGZG22,ChenLPLZY21,WuS0C0Z22}.
In real world, attributed to various connections among objects, learning intact node features from heterogeneous graphs becomes a critical research problem.
Accordingly, this encourages researchers to put more emphases on the Heterogeneous Graph Neural Network (HGNN) for the dexterous manipulation of multi-graph-structured data \cite{Liu2023HGBER,XGWGCT2022,ZhaoWWDWFYL22}.

\begin{figure}[!tbp]
  \centering
  \includegraphics[width=0.7\textwidth]{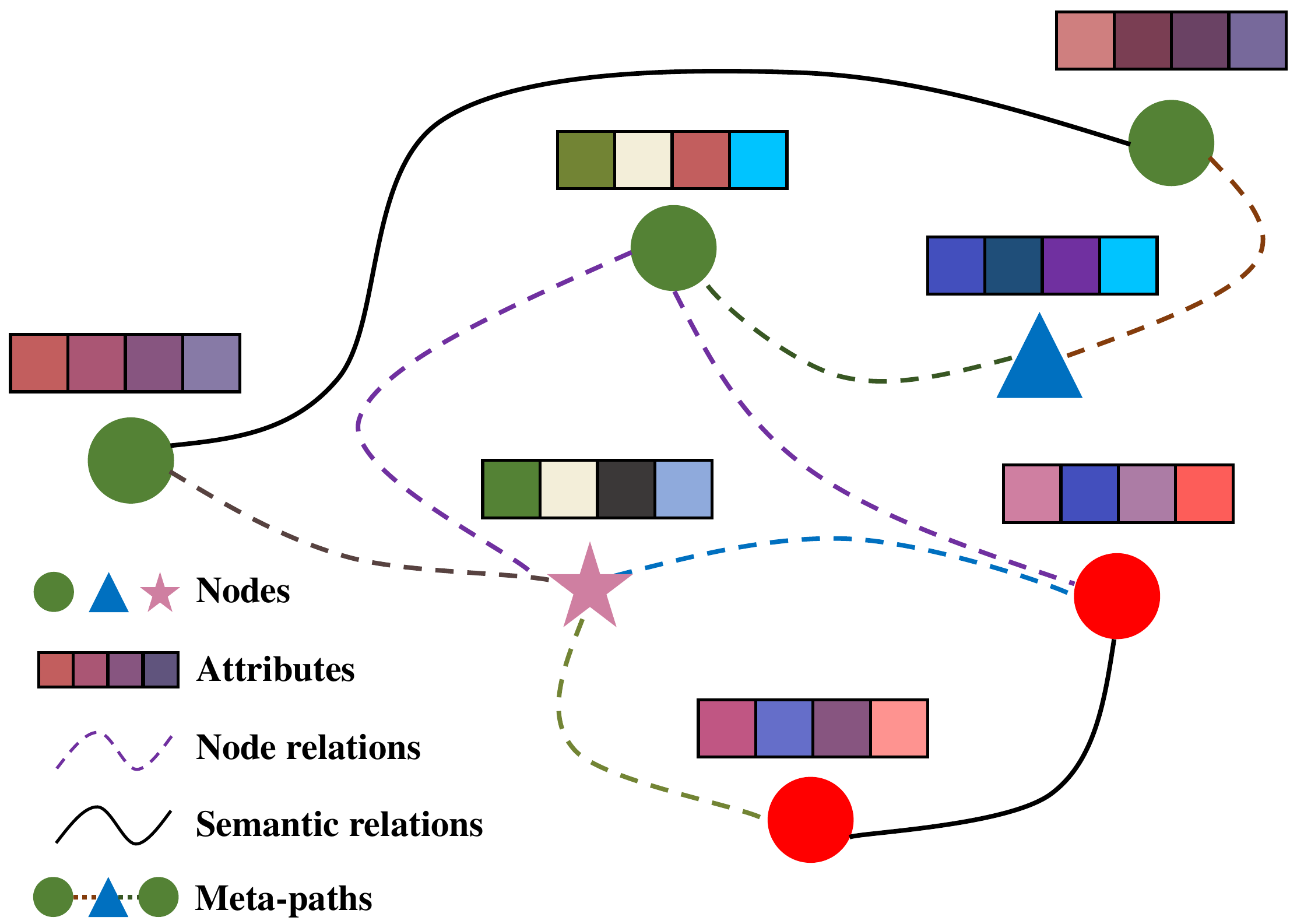}
  \caption{Example of an attributed heterogeneous graph with multiple types of node connections, where different colors of nodes and dotted lines denote various classes of nodes and distinct node relations respectively. Distinct types of nodes are represented by various shapes. The black lines denote the semantic relations that indicate the attributed node homophily.
  }
  \label{AttributedHGraphs}
\end{figure}

Most HGNNs explored node embeddings on the basis of manually designed meta-paths (paths including different dotted lines in Figure \ref{AttributedHGraphs}), and aggregated node information from neighbors discovered by meta-paths \cite{wang2019heterogeneous,wang2022collaborative,qian2022rep2vec}.
For instance, in citation networks, meta-paths w.r.t. authors can be distilled from relations between authors and their papers.
Despite the great success of meta-path-based HGNNs, these models still face the following problems.
On the one hand, the performance of these methods tightly depends on the quality of meta-paths, and models may encounter accuracy decline when some meta-paths are not adequately reliable. 
A simple experiment illustrating this issue is shown in Table \ref{MetaPerformance}, where meta-paths have different semantics in distinct scenarios and are precomputed in prior works.
The performance of GCN varies significantly with distinct defined meta-paths on these widely utilized heterogeneous graph datasets.
It is obvious that GCN gains unfavorable performance with some meta-paths, which results in the undesired classification accuracy when a weighted meta-path is adopted.
This observation indicates that a manually defined meta-path extracted from one or two hops of neighbors is not always satisfactory, and may influence the performance of HGNNs.
To address this drawback, recent works have endeavored to design learnable multi-length meta-paths from short-length meta-paths \cite{YunJKKK19,li2021graphmse,wang2018unsupervised,YuFYHZD22}.
Nevertheless, these methods generally considered a sequential combination of different meta-paths, or only studied refined node relations based on basic meta-paths.
Hence, a more dexterous exploration of multi-length meta-paths should be developed.
On the other hand, most models ignored the natural homophily between attributes of nodes, which is pivotal for finding neighbors in many graph learning methods.
In most cases, long-length meta-paths can connect highly correlated nodes that are remote in the topological space.
Nevertheless, meta-path-based HGNNs often ignored such semantic connections extracted from node attributes, which also benefits the adaptive learning of multi-length meta-paths.

\begin{table}[!tbp]
  \centering
\begin{tabular}{c|cccc}
  \toprule
  Meta-path IDs / Datasets &  ACM & DBLP   & IMDB  & YELP    \\ \midrule
  Meta-path 1  & 0.874     & 0.841      & 0.467     & 0.589   \\
  Meta-path 2  & 0.681     & 0.905      & 0.502     & 0.385 \\
  Meta-path 3  & -         & 0.725      & 0.201     & 0.556 \\
  Weighted fusion & 0.778  & 0.860 & 0.241 & 0.544 \\
\bottomrule
  \end{tabular}
  \caption{Classification performance (Macro-F1) of GCN with different manually defined meta-paths, where weighted fusion indicates that GCN adopts a weighted aggregation of all meta-paths as the adjacency matrix. }
  \label{MetaPerformance}
\end{table}

To better clarify our motivation, 
we utilize Figure \ref{AttributedHGraphs} again to depict an example of an attributed heterogeneous graph.
There are three types of nodes with attributes in the graph, based on which we can propagate node information across different types of nodes via generating meta-paths.
Although some similar nodes are not directly connected, their relations can be captured by the aggregation of meta-paths yielded from other types of nodes and relations.
Consequently, we assume that there exists consistency between the node homophily extracted from node attributes and long-length meta-paths.
Namely, some nodes connected with multi-length meta-paths that are yielded from the fusion of several manually defined short-length meta-paths sometimes share similar node attributes, which motivates us to make full use of node similarity to instruct the design of new long-length meta-paths.
In conclusion, semantic node similarities are generally consistent with meta-path-based node connections,
attributed to which we aim to build a generalized HGNN framework to automatically learn multi-length meta-path fusion under the supervision of node feature homophily.

Therefore, in this paper, we propose an Attributed Multi-Order Graph Convolutional Network (AMOGCN) to learn adaptive multi-order (multi-length) meta-paths from heterogeneous graphs, which is supervised by the semantic node connections extracted from node attribute similarities.
The proposed AMOGCN explores different orders of meta-paths to automatically seek multi-hop node neighbors as much as possible, which considers different permutations of first-order meta-paths dexterously via continuous coefficient matrices.
After that, AMOGCN combines multi-order adjacency matrices with trainable weights.
The aggregation process is instructed by the semantic node homophily, which encourages the model to discover high-order neighbors that have similar attributes.
Eventually, AMOGCN applies a one-layer simplifying GCN to learn high-order node embeddings from the adaptive multi-order meta-path, which is equivalent to the fusion of different multi-layer GCNs. 
In summary, the primary contributions of this paper are listed as follows:
\begin{enumerate}
  \item We develop a multi-order GCN framework addressing attributed heterogeneous graphs via constructing distinct lengths of meta-paths from several manually defined node relations, which is equivalent to the aggregation of multi-hop embeddings captured from various multi-layer GCNs.
  \item The proposed AMOGCN integrates various orders of adjacency matrices adaptively, which is supervised by the semantic information extracted from the homophily between attributed nodes.
  \item We evaluate the proposed model in various heterogeneous graph datasets, and compare it with state-of-the-art works, which reveals that AMOGCN achieves competitive performance in terms of semi-supervised classification tasks.
\end{enumerate}

The rest of this paper is organized as follows. We introduce related works w.r.t. GCNs and heterogeneous graph learning in Section \ref{relatedwork}.
In Section \ref{preliminary}, some basic concepts of heterogeneous graphs are clarified.
We elaborate on the proposed model in Section \ref{TheModel}, and conduct substantial experiments to verify it in Section \ref{experiments}.
Eventually, we conclude our work in Section \ref{conclusion}.

\section{Related Works}\label{relatedwork}
In this section, we review and discuss some related works w.r.t. GCN and heterogeneous graph learning, including computations of high-order adjacency matrices and learnable meta-paths, which are the primary topics of this paper.
\subsection{Graph Convolutional Networks}
GCN has remarkably boosted the learning performance of graph neural networks, each layer of which is defined as
\begin{equation}
  \mathbf{H}^{(l)} = \sigma (\mathbf{A} \mathbf{H}^{(l-1)} \mathbf{W}^{(l)}),
\end{equation}
where $\mathbf{A}$ is the renormalized adjacency matrix and $\mathbf{W}^{(l)}$ is the layer-specific weight matrix.
$\sigma(\cdot)$ is the non-linear activation function.
Wu et al. further simplified GCN via removing the activation function and precomputing a high-order adjacency matrix, which was termed as Simplifying Graph Convolution (SGC) \cite{WuSZFYW19}.
SGC applied the high-order adjacency matrix in a one-layer GCN to approximate the information propagation of multi-layer GCN, i.e.,
\begin{equation}
  \mathbf{H}^{(l)} = \mathbf{A}^{l} \mathbf{X} \mathbf{W},
\end{equation}
where $\mathbf{A}^{l}$ is the $l$-th power of $\mathbf{A}$ and $\mathbf{W}$ is a trainable weight.
Actually, SGC aimed to explore high-order neighbors efficiently with a precomputed adjacency matrix.
A large number of studies have revealed the encouraging performance of GCN and SGC.
Feng et al. designed a cross-feature graph convolution to model the arbitrary-order across node features \cite{FengHZC23}.
Min et al. proposed geometric scattering transformations and residual convolutions to enhance the conventional GCN \cite{MinWW20}.
Xu et al. presented a GCN-based deep feature aggregation model to carry out the high spatial resolution scene classification task \cite{xu2021deep}.
Wu et al. came up with a robust tensor GCN framework to promote the robustness of the model with the multi-view augmentation \cite{WuS0C0Z22}.
Zhu et al. aggregated $k$-step diffusion matrices and proposed a simple spectral graph convolution on the basis of SGC \cite{ZhuK21}.
Wang et al. learned an aggregation of node embeddings from topology space and feature space via a multi-channel GCN \cite{0017ZB0SP20}.
Zhang et al. also leveraged the schema of SGC for attributed graph clustering \cite{0003LLW19}.
Nevertheless, these methods were generally developed for homogeneous graphs, which could not handle more complex multi-relational networks.

\subsection{Heterogeneous Graph Learning}
Owing to the complex relations between objects in the real world, numerous works have paid attention to heterogeneous graph learning with graph neural networks.
A key idea of HGNNs is discovering the consistency and complementarity of different graphs via a joint framework or a fusion strategy.
Yang et al. presented a GCN-based multi-graph fusion method with pseudo-label supervision \cite{YangSJWGY23}.
Sadikaj et al. designed a joint dimensionality reduction algorithm for multi-relational graphs with node attributes, and applied it to the spectral clustering \cite{SadikajVBP21}.
Park et al. proposed an unsupervised node embedding method for attributed multiplex networks, which jointly integrated node features from distinct graphs \cite{ParkK0Y20}.
Zhao et al. presented a heterogeneous graph structure learning framework that fused heterogeneous graphs with a graph neural network \cite{ZhaoWSHSY21}.
A critical factor to obtain satisfactory performance with HGNNs is the design of meta-paths, which significantly affects the quality of learned node embeddings.
Some state-of-the-art works have focused on the automatic learning of meta-paths.
A classical method for meta-path learning is the random walk algorithm \cite{grover2016node2vec,dong2017metapath2vec,shi2018heterogeneous}.
Yun et al. built graph transformer layers to study a soft selection of edge types and generate complex multi-hop neighborhood connections \cite{YunJKKK19}.
Several works have attempted to conduct automatic meta-path selections with HGNNs, which selected and maintained critical meta-paths between nodes \cite{li2021graphmse,wang2018unsupervised}. 
Yu et al. proposed a multiplex heterogeneous GCN which adaptively learned multi-length meta-paths via fusing outputs from different depths of graph convolutional layers \cite{YuFYHZD22}. 
Nonetheless, these automatic graph learning methods seldom considered the permutations of different basic node relations, which are helpful to explore better complementary information from different graphs.
Besides, existing methods for constructing multi-length meta-paths generally lacked suitable supervision.
In light of this, we come up with a multi-order GCN framework that learns adaptive multi-length meta-paths, which is under the supervision of node homophily discovered from node attributes.

\section{Preliminary}\label{preliminary}
In this section, we first introduce some important definitions w.r.t. heterogeneous graphs in this paper.

\begin{definition}\label{HGraph}
\textbf{Heterogeneous Graph} \cite{wang2019heterogeneous}.
A heterogeneous graph is defined as $\mathcal{G} = (\mathcal{V}, \mathcal{E})$, which includes a vertex set $\mathcal{V}$ and an edge set $\mathcal{E}$. 
With the vertex type mapping function $\phi : \mathcal{V} \to \mathcal{O}$ and the edge type mapping function $\phi : \mathcal{E} \to \mathcal{R}$, 
the vertex types and edge types in a heterogeneous graph should satisfy $| \mathcal{O} | + | \mathcal{R} | > 2$.
\end{definition}

\begin{definition}
\textbf{Meta-path} \cite{wang2019heterogeneous}.
A meta-path is defined as the path in the form of $O_{1} \stackrel{R_{1}}{\longrightarrow} O_{2} \stackrel{R_{2}}{\longrightarrow} \cdots \stackrel{R_{h-1}}{\longrightarrow} O_{h}$, which depicts the composite relation $A = R_{1} \circ  R_{2} \cdots  R_{h-1} $ from vertex types $O_{1}$ to $O_{h}$, where $\circ$ is the composition operator on relations.
\end{definition}

Given a heterogeneous graph $\mathcal{G} = (\mathcal{V}, \mathcal{E})$ with distinct types of nodes and relations, we can generate the attributed multi-graph data with various node features $\mathcal{X} = \{ \mathbf{X}_{1}, \mathbf{X}_{2}, \cdots, \mathbf{X}_{|\mathcal{O}|}  \}$ and basic node relations $\mathcal{R} = \{ \mathbf{R}_{1}, \mathbf{R}_{2}, \cdots, \mathbf{R}_{|\mathcal{R} |} \}$.
In this paper, we define the 1-length meta-path as the connection between the same type of nodes, manually constructed from basic node relations among different types of nodes, which corresponds to a first-order adjacency matrix $\mathbf{A}$.
In detail, assume that there is a meta-path manually defined as $O_{1} \stackrel{R_{1}}{\longrightarrow} O_{2} \stackrel{R_{2}}{\longrightarrow} O_{1}$, which includes two types of relations $R_{1}$ and $R_{2}$ and two types of nodes $O_{1}$ and $O_{2}$.
This 1-length directed meta-path can be formulated as a first-order adjacency matrix $\mathbf{A} = \mathbf{R}_{1} \mathbf{R}_{2}$, where adjacency matrices $\mathbf{R}_{1}$ and $\mathbf{R}_{2}$ store the node relations $R_{1}$ and $R_{2}$.
To illustrate, in an online shopping scenario, users $\mathcal{U}$ have two actions for various items $\mathcal{I}$: purchase and collection,
which connect users and items with two different relations.
Herein, a 1-length undirected meta-path $\mathbf{A} $ can be built by $\mathcal{U} \stackrel{purchase}{\longleftrightarrow} \mathcal{I} \stackrel{collection}{\longleftrightarrow}  \mathcal{U}$.
Based on various first-order adjacency matrices, we aim to explore multi-order neighbors for nodes and extract node embeddings automatically via an attributed multi-order GCN.
The primary mathematical notations used in this paper and their explanations are listed in Table \ref{Notations}.

\begin{table}[!tbp]
  \center
  \begin{tabular}{l|l}
  \toprule
   Notations &       Explanations  \\
   \midrule
   $\mathbf{X}$               & Node attributes.  \\
   $\mathbf{R}_{i}$             & The $i$-th node relation matrix.  \\
   $\mathbf{n}_{i}, \mathbf{m}_{i}$    & The numbers of different types of nodes.  \\
   $\mathbf{A}_{i}$             & The $i$-th first-order adjacency matrix.  \\
   $\mathbf{A}_{\mathcal{S}}$   & Semantic adjacency matrix. \\
   $\mathbf{A}^{(l)}$          & The arbitrary $l$-th order adjacency matrix.  \\
   $\mathbf{A}^{(t, l)}$          & The $t$-th $l$-th order adjacency matrix.  \\
   $\bm{\alpha}^{(t,l)}$   & Coefficient matrix for the $t$-th $l$-th order meta-path. \\
   $L$          & Maximum order of high-order adjacency matrices.  \\
   $\mathbf{W}$          & Trainable weight matrix of SGC.  \\
   $\mathcal{A}$          & Multi-order adjacency matrix.  \\
   $\mathbf{H}^{(t, l)}$          & The $t$-th $l$-th order node embeddings.  \\
   $\mathbf{Z}$          & Multi-order node embeddings.  \\
   $\beta^{(t, l)}$   & Weight for the $t$-th $l$-th order adjacency matrix. \\
   $\mathbf{Y}$          & Ground truth.  \\
  \bottomrule
  \end{tabular}
  \caption{A summary of primary notations in this paper.}\label{Notations}
\end{table}

\section{The Proposed Model}\label{TheModel}
\begin{figure}[!tbp]
  \centering
  \includegraphics[width=\textwidth]{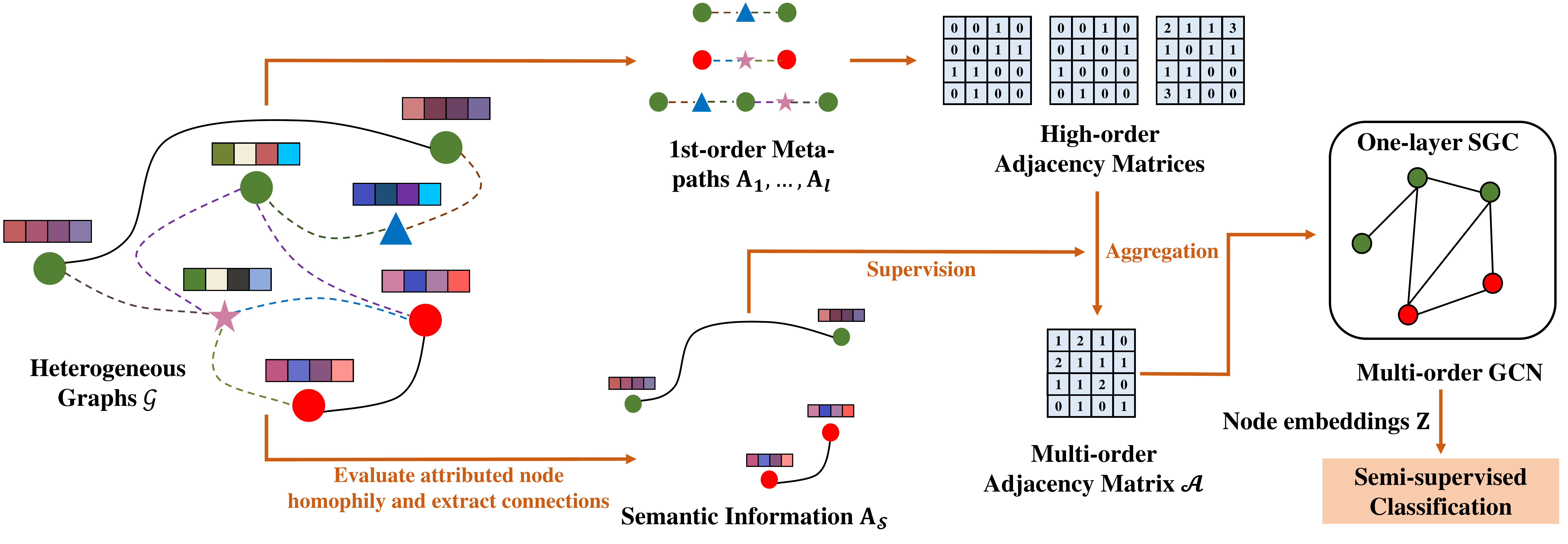}
  \caption{The framework of the proposed AMOGCN. The model first builds first-order meta-paths from heterogeneous graphs, based on which different orders of adjacency matrices are constructed. After that, the model learns a fused multi-order adjacency matrix from high-order meta-paths, supervised by semantic information extracted from node attributes.
  Eventually, the proposed model conducts simplifying graph convolution with the learned multi-order adjacency matrix.
  }
  \label{Framework}
\end{figure}

In this section, we illustrate the proposed AMOGCN which copes with attributed heterogeneous graphs.
Figure \ref{Framework} elaborates on the proposed method.
For the sake of sustained explorations of different lengths of meta-paths and their fusion, we calculate adaptive high-order adjacency matrices from extracted first-order meta-paths.
In addition, we measure the node similarities from attributes and estimate the semantic node connections from node homophily, which are adopted to supervise the fusion of distinct orders of adjacency matrices.
Because the intact multi-order adjacency matrix involves multi-hop neighbor information, we apply a simplifying graph convolutional network that computes node embeddings from high-order neighborhood structures rather than building multiple graph convolutional layers, which saves the training time and corresponds to the neighborhood propagation of the multi-layer GCN.
In the following contents, we introduce and analyze the proposed model via solving the following research problems:

\textbf{(RP1)}. How to construct multi-length meta-paths dexterously via learning multiple high-order adjacency matrices and their fusion?

\textbf{(RP2)}. How to build a multi-order GCN that efficiently propagates node information over learned multi-length meta-paths?

\textbf{(RP3)}. How to adopt node attributes to supervise the fusion of high-order adjacency matrices?

\begin{figure}[!tbp]
  \centering
  \includegraphics[width=0.85\textwidth]{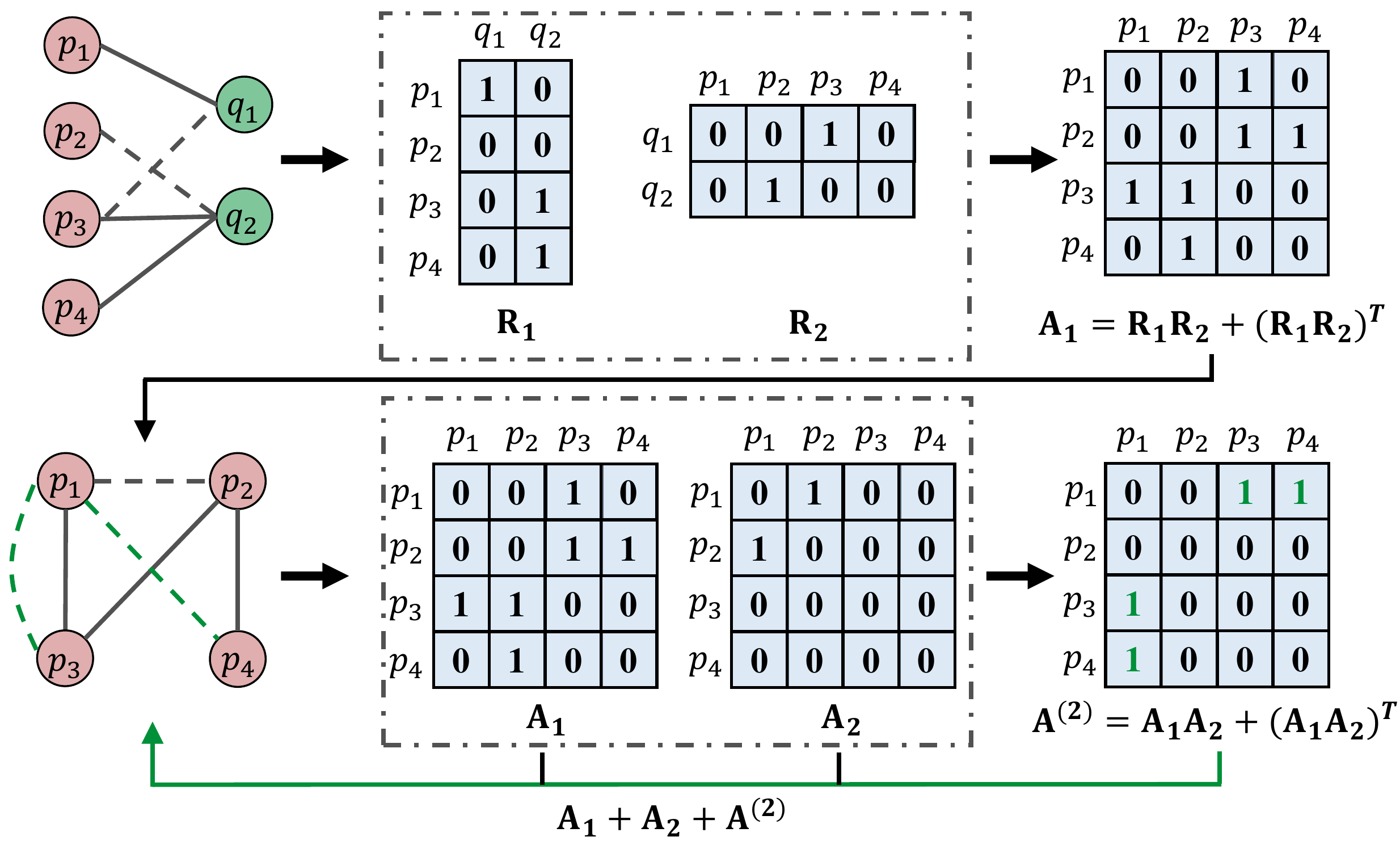}
  \caption{Example of the multi-order meta-path. In this figure, a first-order meta-path $\mathbf{A}_{1}$ is constructed from basic node relations.
  On the basis of 2 different first-order meta-paths $\mathbf{A}_{1}$ and $\mathbf{A}_{2}$, a new second-order meta-path $\mathbf{A}^{(2)}$ is calculated, which generates two new relations between nodes (green dotted lines).
  Fusing all first-order and second-order meta-paths, the new multi-order meta-path is achieved.
  }
  \label{MultiOrderMP}
\end{figure}

\subsection{Multi-order Graph Convolutional Network}
In pursuit of exploring different lengths of meta-paths based on the existing node relations or manually defined meta-paths, we design a multi-order GCN to flexibly study node embeddings with various orders of adjacency matrices, which correspond to multi-length meta-paths.
First, we provide a concrete example to illustrate the meaning of multi-order adjacency matrices.

\textbf{Example.} 
Figure \ref{MultiOrderMP} provides a heterogeneous graph which contains two types of nodes $p$ and $q$, and two types of node relations.
Different node relations can be converted to two adjacency matrices $\mathbf{R}_{1}$ and $\mathbf{R}_{2}$.
For instance, $[\mathbf{R}_{1}]_{ij}$ records the connection from node $p_{i}$ to node $q_{j}$.
We can get an undirected first-order meta-path via $\mathbf{A}_{1} = \mathbf{R}_{1} \mathbf{R}_{2} + (\mathbf{R}_{1} \mathbf{R}_{2})^{T}$.
Namely, $\mathbf{A}_{1}$ stores the undirected meta-path $p \stackrel{R_{1}}{\longleftrightarrow} q \stackrel{R_{2}}{\longleftrightarrow} p$.
Here we temporarily ignore the self-connections of nodes.
Assuming that there is another first-order meta-path $\mathbf{A}_{2}$ yielded from other node relations, we can achieve a new undirected second-order meta-path via $\mathbf{A}^{(2)} = \mathbf{A}_{1}\mathbf{A}_{2} + (\mathbf{A}_{1}\mathbf{A}_{2})^{T}$,
which produces new second-order neighborhood connections.
Accordingly, we can obtain a new multi-order graph with $\mathcal{A} = \mathbf{A}_{1} + \mathbf{A}_{2} + \mathbf{A}^{(2)}$, which consists of the connections to the first-order and the second-order neighbors.
If more other first-order meta-paths are considered, we can generate more different high-order meta-paths.
Thus, we should design a more generalized framework to take full consideration of different permutations of low-order meta-paths, which explores potential multi-hop neighbors more completely.

More generally, an $l$-length ($l$-th order) meta-path can be formulated as a combination of distinct first-order adjacency matrices.
For example, a simple $l$-length meta-path can be represented by a sequential adjacency matrix multiplication, i.e.,
\begin{equation}\label{mo_mx1}
  \mathbf{A}^{(l)} = \mathbf{A}_{1} \mathbf{A}_{2} \cdots \mathbf{A}_{l},
\end{equation}
where $\mathbf{A}_{1} \in \mathbb{R}^{n_{1} \times n_{2}}$, $\mathbf{A}_{2} \in \mathbb{R}^{n_{2} \times n_{3}}$, $\cdots$, $\mathbf{A}_{l} \in \mathbb{R}^{n_{l} \times n_{l+1}}$.
Equation \eqref{mo_mx1} considers the connections to the $l$-hop neighbors.
Actually, $\mathbf{A}^{(l)}$ can be attained by arbitrary permutations of the first-order adjacency matrix multiplications, as long as any $\mathbf{A}_{i} \in \mathbb{R}^{n_{i} \times m_{i}}$ and $\mathbf{A}_{i+1} \in \mathbb{R}^{n_{i+1} \times m_{i+1}}$ satisfy $m_{i} = n_{i+1}$.
In order to consider a more flexible fusion of various manually defined meta-paths, we transform the $l$-th order adjacency matrix in Equation \eqref{mo_mx1} to
\begin{gather}\label{mo_mx2_0}
  \begin{split}
  \mathbf{A}^{(l)} = &\left( \alpha_{11}^{(l)}\mathbf{A}_{1} + \alpha_{12}^{(l)}\mathbf{A}_{2} + \cdots + \alpha_{1l}^{(l)}\mathbf{A}_{l} \right) \cdot \\
  &\left( \alpha_{21}^{(l)}\mathbf{A}_{1} + \alpha_{22}^{(l)}\mathbf{A}_{2} + \cdots + \alpha_{2l}^{(l)}\mathbf{A}_{l} \right) \cdot \\
  & \cdots \\
  &\left( \alpha_{l1}^{(l)}\mathbf{A}_{1} + \alpha_{l2}^{(l)}\mathbf{A}_{2} + \cdots + \alpha_{ll}^{(l)}\mathbf{A}_{l} \right).
  \end{split}
\end{gather}
When $\alpha_{ij}$ is a binary coefficient and $\sum_{j=1}^{l} \alpha_{ij} = 1$ for any $i = 1, 2, \cdots, l$, Equation \eqref{mo_mx1} formulates a high-order adjacency matrix with various permutations of meta-paths with varying $\alpha_{ij}$.
Here, we rewrite Equation \eqref{mo_mx2_0} as
\begin{equation}\label{mo_mx2}
  \mathbf{A}^{(l)} = Product \left(\bm{\alpha}^{(l)} \left[\mathbf{A}_{1}, \mathbf{A}_{2}, \cdots, \mathbf{A}_{l} \right]^{T} \right),
\end{equation}
where $Product (\cdot)$ is the cumulative multiplication of vector elements and
\begin{gather}\label{alpha_mx}
    \bm{\alpha}^{(l)} =
     \left[
     \begin{matrix}
      \alpha_{11}^{(l)} & \cdots & \alpha_{1l}^{(l)} \\
       \vdots & \ddots & \vdots \\
       \alpha_{l1}^{(l)} & \cdots & \alpha_{ll}^{(l)}
      \end{matrix}
      \right]
\end{gather}
is the binary coefficient matrix that the sum of each row is 1.
In particular, when $\bm{\alpha}^{(l)} = \mathbf{I}$, Equation \eqref{mo_mx2} is exactly a sequential cumulative multiplication of adjacency matrices as Equation \eqref{mo_mx1}.
Herein, we relax the discrete coefficient matrix into a continuous one, and renormalize it by a softmax function, i.e.,
\begin{gather}\label{softmax_alpha}
  \bm{\alpha}^{(l)} = Softmax \left( \bm{\alpha}^{(l)}, dim = 1\right),
\end{gather}
which ensures that $\sum_{j=1}^{l} \alpha_{ij} = 1$.
With a directed $\mathbf{A}^{(l)}$, we can further obtain an $l$-th order undirected adjacency matrix via
\begin{equation}\label{ToUndirected}
  \mathbf{A}^{(l)} = \mathbf{A}^{(l)} + {\mathbf{A}^{(l)}}^{T} - diag \left(\mathbf{A}^{(l)} \right).
\end{equation}

In this paper, we set the maximum order of meta-paths as $L$, where $L$ is the number of the first-order meta-paths.
When $l < L$, we consider different selections of basic meta-paths to generate $\mathbf{A}^{(l)}$.
For example, when $L=3$, we can compute $\mathbf{A}^{(2)}$ with subsets $\{ \mathbf{A}_{1},  \mathbf{A}_{2}\}$, $\{ \mathbf{A}_{1},  \mathbf{A}_{3}\}$ and $\{ \mathbf{A}_{2},  \mathbf{A}_{3}\}$.
Notice that Equation \eqref{mo_mx2} has considered various permutations of the given first-order adjacency matrices implicitly.
For simplicity, we denote the set of all $l$-th order adjacency matrices as $\Psi^{(l)} = \{\mathbf{A}^{(t,l)}\}_{t=1}^{|\Psi^{(l)}|}$,
where $|\Psi^{(l)}| = \frac{L!}{l!(L-l)!}$.

Since we have achieved high-order adjacency matrices that contain multi-hop neighborhood information, we only need to adopt a one-layer GCN to propagate node attributes.
Motivated by SGC \cite{WuSZFYW19}, we remove the non-linear activation function and formulate the $t$-th $l$-th order GCN as
\begin{gather}\label{mo_mx3}
  \begin{split}
  \mathbf{H}^{(t, l)} = &\left( \alpha_{11}^{(t, l)}\mathbf{A}_{1}  + \alpha_{12}^{(t, l)}\mathbf{A}_{2}  +  \cdots  + \alpha_{1l}^{(t, l)}\mathbf{A}_{l} \right) \\
  &\left( \alpha_{21}^{(t, l)}\mathbf{A}_{1} + \alpha_{22}^{(t, l)}\mathbf{A}_{2}  +  \cdots  + \alpha_{2l}^{(t, l)}\mathbf{A}_{l} \right) \\
  & \cdots \\
  &\left( \alpha_{l1}^{(t, l)}\mathbf{A}_{1}  +  \alpha_{l2}^{(t, l)}\mathbf{A}_{2}  + \cdots  + \alpha_{ll}^{(t, l)}\mathbf{A}_{l} \right) \\
  &\mathbf{X} \mathbf{W}^{(t, 1)} \mathbf{W}^{(t, 2)} \cdots \mathbf{W}^{(t, l)},
  \end{split}
\end{gather}
where $\mathbf{A}_{1}, \cdots, \mathbf{A}_{l}$ are selected from $L$ precomputed first-order meta-paths.
Particularly, when $\alpha_{ij}^{(t, l)} = 1$ for all $i = 1, \cdots, l$ and $0$ otherwise,
we obtain an $l$-th order adjacency matrix $\mathbf{A}^{(t, l)} = \mathbf{A}_{j}^{l}$ based on the $j$-th first-order meta-path, i.e., the $l$-th power of $\mathbf{A}_{j}$.
In this case, Equation \eqref{mo_mx3} becomes a standard SGC without non-linear activation functions, i.e.,
\begin{equation}
  \mathbf{H}^{(t, l)} = \mathbf{A}_{j}^{l} \mathbf{X} \mathbf{W}^{(t, 1)} \mathbf{W}^{(t, 2)} \cdots \mathbf{W}^{(t, l)}.
\end{equation}
It can be regarded as the $l$-hop graph information propagation over the topology network denoted by $\mathbf{A}_{j}^{l}$.

Replacing $\mathbf{W}^{(t, 1)} \mathbf{W}^{(t, 2)} \cdots \mathbf{W}^{(t, l)}$ with a shared trainable matrix $\mathbf{W}^{(t, l)}$, we can simplify Equation \eqref{mo_mx3} into
\begin{gather}\label{mo_mx4}
  \mathbf{H}^{(t, l)} = \mathbf{A}^{(t, l)} \mathbf{X} \mathbf{W}^{(t, l)},
\end{gather}
where $\mathbf{A}^{(t, l)}$ denotes the $t$-th $l$-th order adjacency matrix defined in Equation \eqref{mo_mx3}.
Moreover, if different lengths of meta-paths are considered, we have
\begin{gather}\label{mo_mx5}
  \begin{split}
  \mathbf{Z} &= \sum_{l=1}^{L} \sum_{t=1}^{|\Psi^{(l)}|} \beta^{(t, l)} \mathbf{H}^{(t, l)} \\
   &= \sum_{l=1}^{L} \sum_{t=1}^{|\Psi^{(l)}|}  \beta^{(t, l)} \left( \mathbf{A}^{(t, l)} \mathbf{X} \mathbf{W}^{(t, l)} \right),
  \end{split}
\end{gather}
where $\beta^{(t, l)}$ is a trainable parameter and $\sum_{l=1}^{L} \sum_{t=1}^{|\Psi^{(l)}|}  \beta^{(t, l)} = 1$.
We adopt the softmax function to project $\beta^{(t, l)}$ onto the feasible space.
Considering a shared $\mathbf{W}$ for all trainable weight matrices,
Equation \eqref{mo_mx5} leads to 
\begin{gather}\label{mo_mx6}
  \mathbf{Z} = \mathcal{A} \mathbf{X} \mathbf{W},
\end{gather}
where $\mathcal{A}$ is the multi-order adjacency matrix defined as
\begin{gather}\label{mo_adj}
  \mathcal{A} = \sum_{l=1}^{L} \sum_{t=1}^{|\Psi^{(l)}|}  \beta^{(t, l)} \mathbf{A}^{(t, l)}.
\end{gather}
In summary, Equations \eqref{mo_mx2}, \eqref{softmax_alpha}, \eqref{ToUndirected} and \eqref{mo_adj} explain the construction of multi-order adjacency matrix (\textbf{RP1}), on the basis of which we establish the multi-order GCN with Equation \eqref{mo_mx6} (\textbf{RP2}).

\subsection{Semantic Information Supervision}
Owing to the complexity of heterogeneous graphs, some highly similar nodes may be disconnected in the topology space.
Because the aggregation of multi-order meta-paths can describe the node relations more comprehensively, it should also be correlated with node homophily information.
Thus, apart from serving as the input of GCN, node attributes are leveraged to build a semantic adjacency matrix that describes the node homophily to further instruct the learning of multi-order meta-path aggregation (\textbf{RP3}).
The semantic adjacency matrix can be evaluated by some pair-wise distance measurements that compute the spatial distance based on node feature vectors, such as cosine and Gaussian similarities.
We select the top-$k$ similar neighbors for each node to generate the semantic adjacency matrix.
Formulaically, we construct the semantic adjacency matrix via

\begin{equation}\label{knn}
  \mathbf{A}_{\mathcal{S}} = 
  \left\{\begin{matrix}
    1, \; x_{i} \in TopK(x_{j}) \; or \; x_{j} \in TopK(x_{i}),\\
    0, \; otherwise.\\
    \end{matrix}\right.
\end{equation}

With the precomputed $\mathbf{A}_{\mathcal{S}}$, we require that the multi-order adjacency matrix fusion should involve the critical connections estimated by node homophily.
This is because that a high-order neighborhood relation should tend to connect nodes that are remote but similar.
Accordingly, the supervision of semantic information can be measured by the loss function
\begin{gather}\label{recError}
  \mathcal{L}_{rec} \left( \mathcal{A}, \mathbf{A}_{\mathcal{S}} \right)  = - \frac{1}{\kappa} \sum_{i,j} \left[\mathbf{A}_\mathcal{S} \right]_{ij} log \left( \mathcal{A}_{ij} \right),
\end{gather}
where $\kappa$ is the number of non-zero entries in $\mathbf{A}_\mathcal{S}$.
It is noted that the semantic node homophily information only indicates the potential connections between similar attributed nodes, and we should not ignore other types of node relations.
Consequently, we only consider the supervision from existing semantic connections.

Combining both label and semantic information supervision, the loss function of AMOGCN for the semi-supervision classification is 
\begin{equation}\label{totalloss}
\mathcal{L} \left(\mathbf{Z}, \mathbf{Y}, \mathcal{A}, \mathbf{A}_{\mathcal{S}} \right) = \mathcal{L}_{ce} \left(\mathbf{Z}, \mathbf{Y} \right) + \gamma \mathcal{L}_{rec} \left( \mathcal{A}, \mathbf{A}_{\mathcal{S}} \right),
\end{equation}
where $\gamma$ is the trade-off hyperparameter and $\mathcal{L}_{ce}$ is the cross-entropy loss function, i.e.,
\begin{gather}\label{celoss}
  \mathcal{L}_{ce} \left(\mathbf{Z}, \mathbf{Y} \right) = 
   - \sum_{i \in \Omega} \sum_{j=1}^{c} \mathbf{Y}_{ij} \mathrm{ln} \mathbf{Z}_{ij},
\end{gather}
where $\mathbf{Y}$ is the matrix containing the label information from the training set $\Omega$.

\subsection{Training Algorithm}
\begin{algorithm}[!tbp]
  \caption{Attributed Multi-Order Graph Convolutional Network (AMOGCN)}\label{algorithm}
  \KwIn{Node attributes $\mathbf{X} \in\mathbb{R}^{n \times f}$, first-order adjacency matrices $\{\mathbf{A}_{i} \}_{i=1}^{L}$, ground truth $\mathbf{Y} \in\mathbb{R}^{n \times c}$, hyperparameters $\gamma$ and $k$.}
  \KwOut{Node embedding $\mathbf{Z}$.}
  Obtain the semantic adjacency matrix $\mathbf{A}_{\mathcal{S}}$ for supervision\;
  Initialize all coefficient matrices $\bm{\alpha}^{(t, l)}$ randomly\;
  \While{$\mathcal{L} \left(\mathbf{Z}, \mathbf{Y}, \mathcal{A}, \mathbf{A}_{\mathcal{S}} \right)$ does not converge}
  {
    \tcc{Adaptive high-order meta-paths learning.}
    Obtain different orders of adjacency matrices $\mathbf{A}^{(t, l)}$ that contain various lengths of meta-paths with Equations \eqref{mo_mx2}, \eqref{softmax_alpha} and \eqref{ToUndirected}\;
    \tcc{Multi-order meta-path aggregation.}
    Compute the multi-order adjacency matrix $\mathcal{A}$ that fuses distinct lengths of meta-paths with Equation \eqref{mo_adj}\;
    \tcc{Multi-order feature learning with SGC.}
    Compute outputs of the multi-order GCN $\mathbf{Z}$ with Equation \eqref{mo_mx6}\;
    Calculate the loss $\mathcal{L} \left(\mathbf{Z}, \mathbf{Y}, \mathcal{A}, \mathbf{A}_{\mathcal{S}} \right)$ with Equation \eqref{totalloss}\;
    Update trainable GCN weight $\mathbf{W}$, coefficient matrix $\bm{\alpha}$ and all weights $\beta^{(t,l)}$ with back propagation\;
  }
  \Return{Multi-order node embedding $\mathbf{Z}$.}
  \end{algorithm}
We elaborate on the detailed training procedure of AMOGCN in Algorithm \ref{algorithm}.
Overall, the forward propagation of the network consists of three steps.
(1) Calculate different orders of meta-paths adaptively with trainable coefficients (Line 4).
(2) Obtain a multi-order meta-path fusion with learnable weights (Line 5).
(3) Conduct SGC with the precomputed multi-order meta-path (Line 6).
Notice that we may get some repeated or closed-loop paths if the lengths of meta-paths are long.
Thus, the fusion of exorbitant lengths of meta-paths often yields redundant node relations and massive self-connections, which result in the meaningless high-order adjacency matrix. 
On the other hand, because the proposed AMOGCN corresponds to the fusion of multi-layer GCNs with various first-order adjacency matrices,
the multi-length meta-paths with repeated types of node connections in AMOGCN are equivalent to the homogeneous deep-layer GCN.
This often leads to notorious performance owing to the oversmoothing issue when a GCN is too deep.
Hence, we only consider the maximum order of adjacency matrices as the number of the first-order meta-paths to reduce unnecessary computations.
Actually, because we adopt SGC to propagate node information over the multi-order adjacency matrix, the time cost is much lower than a multi-layer GCN.

\subsection{Differences to Existing Works}
AMOGCN considers the automatic learning of the multi-order adjacency matrix, which is also investigated by some prior studies as we discussed in related works.
Our model differs from existing methods (e.g., SGC \cite{WuSZFYW19}, MHGCN \cite{YuFYHZD22}) in the following aspects:

(1) \textbf{\textit{AMOGCN is a more generalized framework exploring multi-order or multi-length meta-paths.}} For example, MHGCN and SGC can be regarded as special cases of AMOGCN. It is exactly SGC when a high-order adjacency matrix generated from a specific first-order meta-path is adopted. MHGCN is a multi-layer network that performs the weighted sum of outputs from distinct layers, which is equivalent to the one-layer AMOGCN that only considers sequential permutations of the first-order meta-paths.

(2) \textbf{\textit{The aggregation of different orders of adjacency matrices is considered, which attempts to cover remote and nearby neighborhood connections at the same time.}}
The adaptive fusion of different meta-paths enables the final multi-order adjacency matrix to capture more consistency and complementary information from different types of node relations.

(3) \textbf{\textit{We make full use of node attributes in heterogeneous graphs.}} Most prior methods only utilized node attributes as the input of GCNs. In our work, apart from serving as initial embeddings of graph convolutions, node attributes are applied to estimate the semantic node relations containing node homophily, which instruct the learning of the multi-order adjacency matrix aggregation.

\section{Experimental Analysis}\label{experiments}
In this section, we evaluate the proposed AMOGCN with comprehensive experiments. The proposed model is implemented with PyTorch and runs on a machine with AMD R9-5900HS CPU, RTX 3080 16G GPU and 32G RAM.

\subsection{Experimental Setup}

\textbf{Compared Baselines}.
In this paper, we compare the proposed model with classical and state-of-the-art baselines, including GCN \cite{KipfW17}, SGC \cite{WuSZFYW19}, HAN \cite{wang2019heterogeneous}, DGI \cite{VelickovicFHLBH19}, DMGI \cite{ParkK0Y20}, IGNN \cite{NEURIPS2020_8b5c8441}, SSDCM \cite{MitraVSG0R21} and MHGCN \cite{YuFYHZD22}.
Table \ref{Methods} lists the characteristics of these models.
The compared methods can be divided into two types of models: homogeneous-network-based frameworks (GCN, SGC and DGI) and heterogeneous-network-based frameworks (HAN, DMGI, IGNN, SSDCM and MHGCN). 
For homogeneous-network-based models, we test the model with average weighted meta-paths.
Some important related works that also explore learnable meta-paths or multi-order meta-paths are compared in our experiments (SGC and MHGCN).
In detail, the description and code links of these methods are given below.

\begin{table}[!tbp]
  \centering
\begin{tabular}{lcccc}
  \toprule
  Methods  & Attr.         & Heter.        & Learnable.  & Multi-order.      \\ \midrule
  GCN \cite{KipfW17}      & $\checkmark$  & $\times$      & $\times$     & $\times$   \\
  SGC \cite{WuSZFYW19}     & $\checkmark$  & $\times$      & $\times$     & $\checkmark$ \\
  DGI \cite{VelickovicFHLBH19}     & $\checkmark$  & $\times$      & $\times$     & $\times$   \\
  HAN \cite{wang2019heterogeneous}     & $\checkmark$  & $\checkmark$  & $\times$     & $\times$   \\
  DMGI \cite{ParkK0Y20}    & $\checkmark$  & $\checkmark$  & $\times$     & $\times$    \\
  IGNN \cite{NEURIPS2020_8b5c8441}    & $\checkmark$  &$\checkmark$    & $\times$     & $\times$   \\
  SSDCM \cite{MitraVSG0R21}   & $\checkmark$  &$\checkmark$    & $\times$     & $\times$   \\
  MHGCN \cite{YuFYHZD22}   & $\checkmark$  & $\checkmark$  & $\checkmark$ & $\checkmark$\\
  AMOGCN (Ours)  & $\checkmark$  & $\checkmark$  & $\checkmark$ & $\checkmark$\\
\bottomrule
  \end{tabular}
  \caption{Properties of different models (Attr.: Using attributes, Heter.: For heterogeneous graphs, Learnable.: Learnable meta-paths, Multi-order.: Multi-order meta-paths).}
  \label{Methods}
\end{table}

\begin{itemize}
	\item \textbf{GCN}\footnote{https://github.com/tkipf/gcn} \cite{KipfW17} is a semi-supervised homogeneous graph convolutional network which obtains node embeddings by aggregating message from local neighborhood structures.
	
	\item \textbf{SGC}\footnote{https://github.com/Tiiiger/SGC} \cite{WuSZFYW19} is a simplified version of GCN framework, which only employs the product of high-order adjacency matrices and attribute matrix, removing non-linear transformation for the semi-supervised classification tasks.
	
	\item \textbf{HAN}\footnote{https://github.com/Jhy1993/HAN} \cite{wang2019heterogeneous} explores the node-level and semantic-level attention on multiplex networks to learn the importance of nodes and meta-paths, thereby generating node representations in a hierarchical manner.
	
	\item \textbf{DGI}\footnote{https://github.com/PetarV-/DGI} \cite{VelickovicFHLBH19} is an unsupervised graph learning representation approach which maximizes mutual information between the graph-level summary embeddings and the local patches to obtain global graph structures.
	
	\item \textbf{DMGI}\footnote{https://github.com/pcy1302/DMGI} \cite{ParkK0Y20} is an unsupervised attributed multiplex network which jointly integrates the node embeddings from multiple relations to learn high-quality representations through a consensus regularization framework and a universal discriminator for downstream tasks.
	
	\item \textbf{IGNN}\footnote{https://github.com/SwiftieH/IGNN} \cite{NEURIPS2020_8b5c8441} is a graph learning framework which employs a fixed-point equilibrium equation and the Perron-Frobenius theory to iterate graph convolutional aggregation until converging for node classification tasks.
	
	\item \textbf{SSDCM}\footnote{https://github.com/anasuamitra/ssdcm} \cite{MitraVSG0R21} is a semi-supervised framework for representation learning which aims to maximize the mutual information between local and contextualized global graph summaries and employs the cross-layer links to impose the regularization of the node embeddings.
	
	\item \textbf{MHGCN}\footnote{https://github.com/NSSSJSS/MHGCN} \cite{YuFYHZD22} automatically learns the useful relation-aware topological structural signals by the multiplex relation aggregation and a multi-layer graph convolution for graph representation learning tasks.
\end{itemize}

\begin{table}[!tbp]
  \centering
  \resizebox{\textwidth}{!}{
\begin{tabular}{cccccc}
  \toprule
  Datasets    &   \# Nodes & \# Attributes   & Node types & Manually defined meta-paths       & \# Classes         \\ \midrule
  ACM      &     8,916     & 1,870            & Paper (P) / Author (A) / Subject (S)  & PAP, PSP      & 3        \\
  DBLP   &   27,194   & 334             & Paper (P) / Author (A) / Conference (C) / Term (T)     & APA, APCPA, APTPA   & 4      \\
  IMDB   & 12,722 & 1,232            & Movie (M) / Actor (A) / Director (D) / Year (Y)          & MAM, MDM, MYM   &   3        \\
  YELP   & 3,913  & 82              & Business (B) / User (U) / Service (S) / Level (L)   & BUB, BLB, BSB     &  3        \\\bottomrule
  \end{tabular}}
  \caption{Statistics of tested heterogeneous graph datasets.}
  \label{Datasets}
\end{table}

\textbf{Datasets}.
Four publicly available datasets are adopted to evaluate the performance of compared methods, i.e., ACM, DBLP, IMDB and YELP.
All of these datasets contain heterogeneous graphs with attributed nodes, and some manually defined meta-paths are precomputed from multiple node relations.
Detailed descriptions are given as follows.
\begin{itemize}
	\item \textbf{ACM} is a citation network dataset which contains 3,025 nodes divided into three types of nodes i.e., paper, author and subject.
	All nodes are leveraged to construct citation networks, paper content, and other data integration studies.
	We employ the meta-path set \{PAP, PSP\} for experiments.
	
	\item \textbf{DBLP} is extracted from the DBLP citation network website with each node having 334 attributes.
	All the nodes are classified into four categories, i.e., author, paper, term and conference.
	The meta-path set \{APA, APCPA, APTPA\} are employed to conduct experiments.
	
	\item \textbf{IMDB} is a movie dataset containing four types of nodes i.e., movie, actor, director and year.
	Nodes are divided into three classes i.e., action, comedy, drama according to the movie genre.
	Movie features correspond to elements of a bag-of-words representation of plots.
	We employ the meta-path set \{MAM, MDM, MYM\} to perform experiments.
	
	\item \textbf{YELP} is a subset derived from the merchant review website with four types of nodes, i.e., business, user, service and level.
  We generate the meta-path set \{BUB, BLB, BSB\} to conduct experiments.
\end{itemize}

More statistics of these datasets are summarized in Table \ref{Datasets}.

\textbf{Performance Evaluation}.
We evaluate the classification performance via the widely used Macro-F1 score and Micro-F1 score.
All methods are repeated 5 times and we record the average performance.
We randomly split all datasets into different ratios of the training sets (20\%/40\%/60\%), validation sets (10\%) and test sets (10\%) for classification performance evaluation.

\subsection{Experimental Results}

\begin{table*}[!tbp]
  \centering
  \resizebox{\textwidth}{!}{\begin{tabular}{c|c|c|ccc|ccccc|c}
    \toprule
  Datasets                 & Training              & Metrics   & GCN & SGC & DGI  & HAN & DMGI & IGNN & SSDCM & MHGCN & AMOGCN \\\midrule 
  \multirow{7}{*}{ACM}     & \multirow{2}{*}{20\%} & Macro-F1 & 0.786  & 0.675  & 0.224  & 0.915  & 0.867 & 0.809  & 0.877 & \underline{0.889}  & \textbf{0.924}            \\
                           &                       & Micro-F1 & 0.788  & 0.702  & 0.369  & 0.914  & 0.868 & 0.795  & 0.876 & \underline{0.891}  & \textbf{0.925}            \\ \cmidrule{2-12}
                           & \multirow{2}{*}{40\%} & Macro-F1  & 0.751  & 0.672  & 0.227  & 0.910  & 0.871 & 0.872 & 0.881 & \underline{0.911}  & \textbf{0.943}            \\
                           &                       & Micro-F1  & 0.758  & 0.699  & 0.377  & 0.911  & 0.868 & 0.871& 0.883 & \underline{0.918}  & \textbf{0.944}           \\\cmidrule{2-12}
                           & \multirow{2}{*}{60\%} & Macro-F1  & 0.764 & 0.689  & 0.241  & 0.892  & 0.912 &0.903 & 0.886 & \underline{0.937}  & \textbf{0.948}          \\
                           &                       & Micro-F1  & 0.755  & 0.649  & 0.375  & 0.891  & 0.909 & 0.904& 0.888 & \underline{0.924}  & \textbf{0.947}      \\
                           \midrule 
  \multirow{7}{*}{DBLP}    & \multirow{2}{*}{20\%} & Macro-F1  & 0.901  & 0.873  & 0.243  & 0.893  & 0.657 & 0.891& 0.894 & \underline{0.909}  & \textbf{0.924}         \\
                           &                       & Micro-F1  & 0.916  & 0.904  & 0.376  & 0.904  & 0.711 & 0.902& 0.899 & \underline{0.921}  & \textbf{0.943}        \\\cmidrule{2-12}
                           & \multirow{2}{*}{40\%} & Macro-F1 & 0.895  & 0.816  & 0.241  & 0.900  & 0.714 &0.890 & \underline{0.902} & 0.895  & \textbf{0.916}       \\
                           &                       & Micro-F1  & 0.906  & 0.850  & 0.366  & 0.904  & 0.773 &0.884 & \underline{0.906} & 0.904  & \textbf{0.921}     \\\cmidrule{2-12}
                           & \multirow{2}{*}{60\%} & Macro-F1 & 0.902  & 0.850  & 0.368  & 0.907  & 0.721 &0.895 & 0.909 & \underline{0.922}  & \textbf{0.929}         \\
                           &                       & Micro-F1  & 0.909  & 0.870  & 0.386  & 0.911  & 0.787 &0.904 & 0.911 & \textbf{0.936}  & \underline{0.933}     \\
                           \midrule 
  \multirow{7}{*}{IMDB}    & \multirow{2}{*}{20\%} & Macro-F1  & 0.243  & 0.270  & 0.263  & 0.498  & 0.353 & 0.403 & 0.494 & \textbf{0.505}  & \underline{0.502}         \\
                           &                       & Micro-F1 & 0.554  & 0.548  & 0.552  & 0.549  & 0.573 & 0.500& 0.591 & \underline{0.642}  & \textbf{0.651}           \\\cmidrule{2-12}
                           & \multirow{2}{*}{40\%} & Macro-F1  & 0.247  & 0.285  & 0.264  & 0.520  & 0.382 & 0.503 & 0.521 & \underline{0.523}  & \textbf{0.533}          \\
                           &                       & Micro-F1  & 0.544  & 0.550  & 0.541  & 0.540  & 0.590 & 0.575 & 0.592 & \underline{0.612}  & \textbf{0.628}       \\\cmidrule{2-12}
                           & \multirow{2}{*}{60\%} & Macro-F1  & 0.287  & 0.301  & 0.271  & 0.542  & 0.397 & 0.516 & \underline{0.549} & 0.531  & \textbf{0.551}      \\
                           &                       & Micro-F1 & 0.504  & 0.563  & 0.565  & 0.569  & 0.610  & 0.588 & 0.601 & \underline{0.615}  & \textbf{0.678}           \\
                           \midrule 
  \multirow{7}{*}{YELP} & \multirow{2}{*}{20\%} & Macro-F1  & 0.520  & 0.519  & 0.503  & 0.483  & 0.516  & \underline{0.642} & 0.527 & 0.546  & \textbf{0.655}        \\
                           &                       & Micro-F1  & 0.674  & 0.674  & 0.683  & 0.489  & 0.699 & \underline{0.712} & 0.702 & 0.707  & \textbf{0.724}         \\\cmidrule{2-12}
                           & \multirow{2}{*}{40\%} & Macro-F1  & 0.530  & 0.535  & 0.542  & 0.458  & 0.534 & \underline{0.645} & 0.542 & 0.553  & \textbf{0.660}           \\
                           &                       & Micro-F1  & 0.713  & 0.720  & 0.719  & 0.552  & 0.709 & \underline{0.711} & 0.707 & 0.697  & \textbf{0.724}          \\\cmidrule{2-12}
                           & \multirow{2}{*}{60\%} & Macro-F1 & 0.580  & 0.567  & 0.543  & 0.439  & 0.546 & \underline{0.671} & 0.587 & 0.598  & \textbf{0.692}         \\
                           &                       & Micro-F1 & 0.736  & 0.782  & 0.723  & 0.529  & 0.721 & 0.624 & 0.722 & \underline{0.739}  & \textbf{0.751}         \\
                           \bottomrule  
  \end{tabular}}
  \caption{Node classification performance with various percentages of training samples.}\label{ClassificationPerformance}
  \end{table*}

\begin{figure*}[!tbp]
  \centering
  \includegraphics[width=\textwidth]{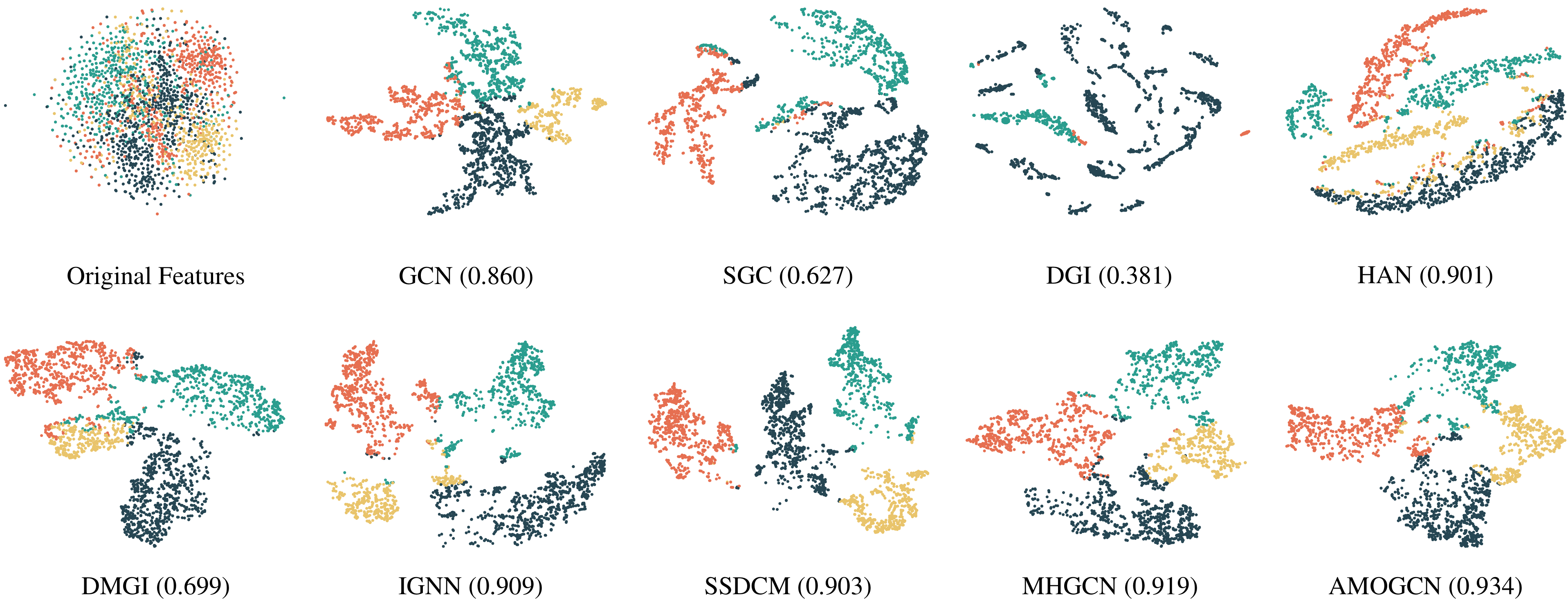}
  \caption{T-SNE Visualization and Micro-F1 values of compared methods on DBLP dataset with 20\% training samples.
  }
  \label{VisualizationPerformance}
\end{figure*}

\textbf{Classification Results}.
In this paper, we run AMOGCN with a learning rate fixed as 0.01, and the Adam optimizer is adopted.
Table \ref{ClassificationPerformance} exhibits the classification performance with varying training ratios on different datasets.
In most cases, HGNNs are able to gain remarkable performance with a few node labels.
The experimental comparisons reveal that AMOGCN achieves competitive classification performance compared with state-of-the-art baselines.
In general, methods for homogeneous graphs, especially SGC and DGI, perform poorly on node embedding learning.
It is noted that SGC precomputes a high-order adjacency matrix from a weighted sum of existing meta-paths.
The undesired classification accuracy indicates that the learnable permutations of first-order meta-paths are necessary when computing the high-order adjacency matrix.
Among all heterogeneous models, MHGCN and the proposed AMOGCN behave the best, revealing that adaptive multi-order meta-paths are beneficial to the graph embedding learning.
We also visualize the classification results of various models on DBLP dataset, as shown in Figure \ref{VisualizationPerformance}.
It can be observed that most compared methods generally succeed in learning separable node features, while several compared approaches have some significant mixed nodes belonging to different classes.
IGNN, SSDCM, MHGCN and the proposed AMOGCN obtain higher classification accuracy with a stronger ability to get more distinguishable node clusters.
In general, these methods gain competitive performance with few training samples (e.g., 20\% training data).
AMOGCN performs even better and the intra-class correlations in each cluster are closer, which may be attributed to the supervision of semantic information extracted from node homophily.
These observations verify the superior node representation learning ability of the proposed AMOGCN.

\begin{figure}[!tbp]
  \centering
  \includegraphics[width=0.7\textwidth]{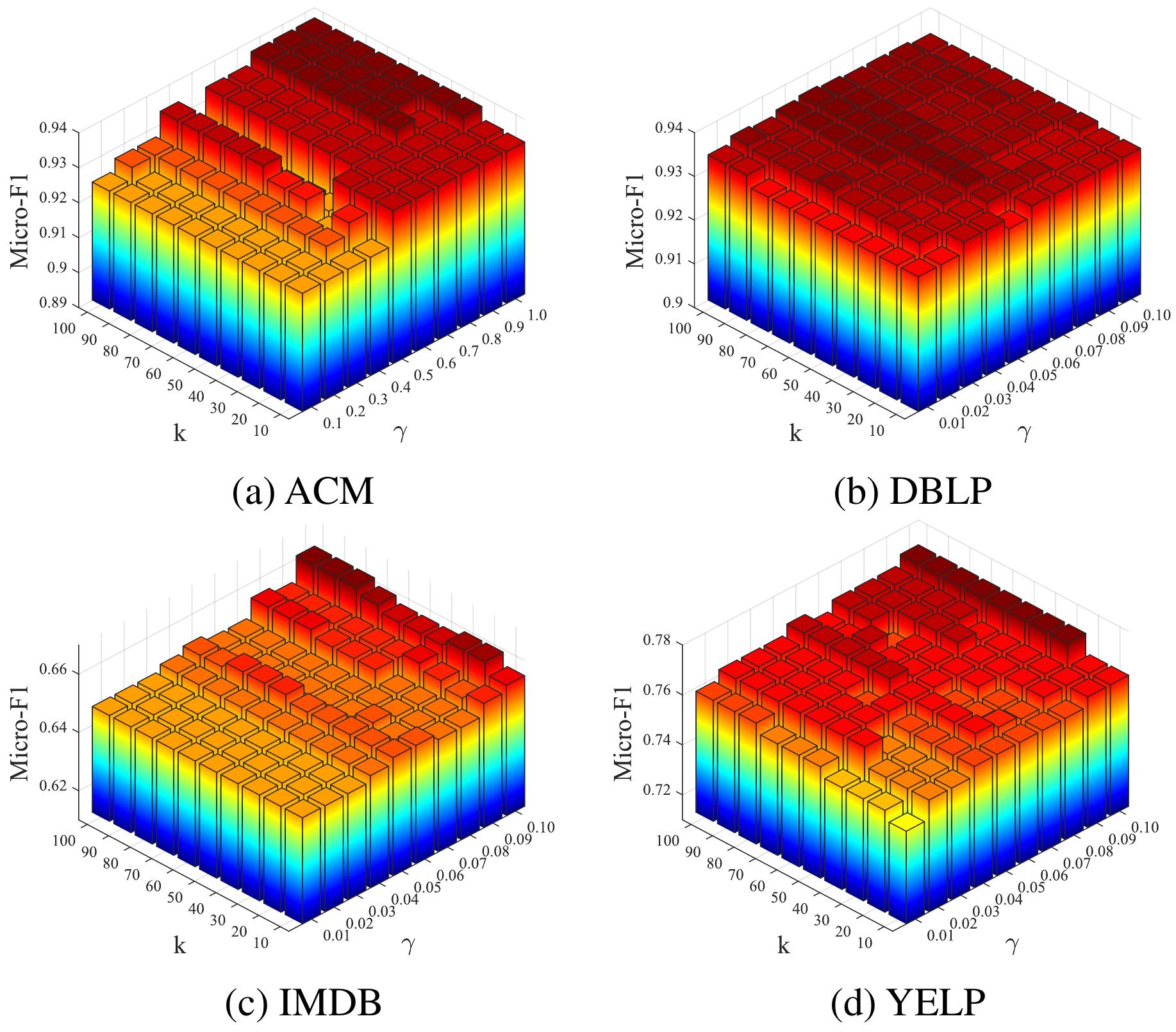}
  \caption{Parameter sensitivity w.r.t. hyperparameters $k$ and $\gamma$ in AMOGCN.
  }
  \label{ParameterSensitivity}
\end{figure}

\textbf{Parameter Sensitivity}.
We analyze the impact of hyperparameters $k$ and $\gamma$ for semantic information supervision in Figure \ref{ParameterSensitivity}, from which we can obtain the following discoveries.
It is obvious that a higher $\gamma$ promotes the classification performance on all datasets, and the model reaches the best accuracy when $\gamma$ is large enough.
The minimal optimal $\gamma$ varies on different datasets, but generally it is larger than $0.05$.
With a fixed $\gamma$, a larger $k$ also often leads to higher accuracy, and the model generally can not obtain more performance benefits when $k > 50$.
This is because that an attributed node only has a limited number of semantic neighbors in most cases.
These observations indicate that the semantic information extracted from adequate numbers of neighborhood nodes facilitates the model to learn more tailored fusion of multi-order adjacency matrices, which endows the model with a more powerful ability to study node embeddings from suitable high-order neighbors.

\textbf{Ablation Study}.
We also conduct the ablation study to verify the effectiveness of semantic supervision, as recorded in Table \ref{Ablation}.
The experimental results show that the supervision of semantic information works on different datasets, which validate that the aggregation of distinct orders of adjacency matrices can be effectively guided by the node homophily.
It is clear that the node relations extracted from node similarities promote the classification accuracy, which is more significant on IMDB and YELP datasets.
Therefore, this also validates that some attributed nodes connected by long-length meta-paths are similar in the feature space.

\begin{table}[!tbp]
  \centering
\begin{tabular}{ccccc}
  \toprule
  Methods  & ACM         & DBLP        & IMDB   & YELP     \\ \midrule
  AMOGCN w/o Sema.     & 0.912      & 0.933      & 0.644     & 0.709   \\
  AMOGCN  & \textbf{0.925}  & \textbf{0.943} & \textbf{0.651} & \textbf{0.724} \\
\bottomrule
  \end{tabular}
  \caption{Ablation study of AMOGCN with 20\% training samples, evaluated by Micro-F1 values (AMOGCN w/o Sem.: AMOGCN without semantic information supervision).}
  \label{Ablation}
\end{table}

\begin{figure}[!tbp]
  \centering
  \includegraphics[width=0.8\textwidth]{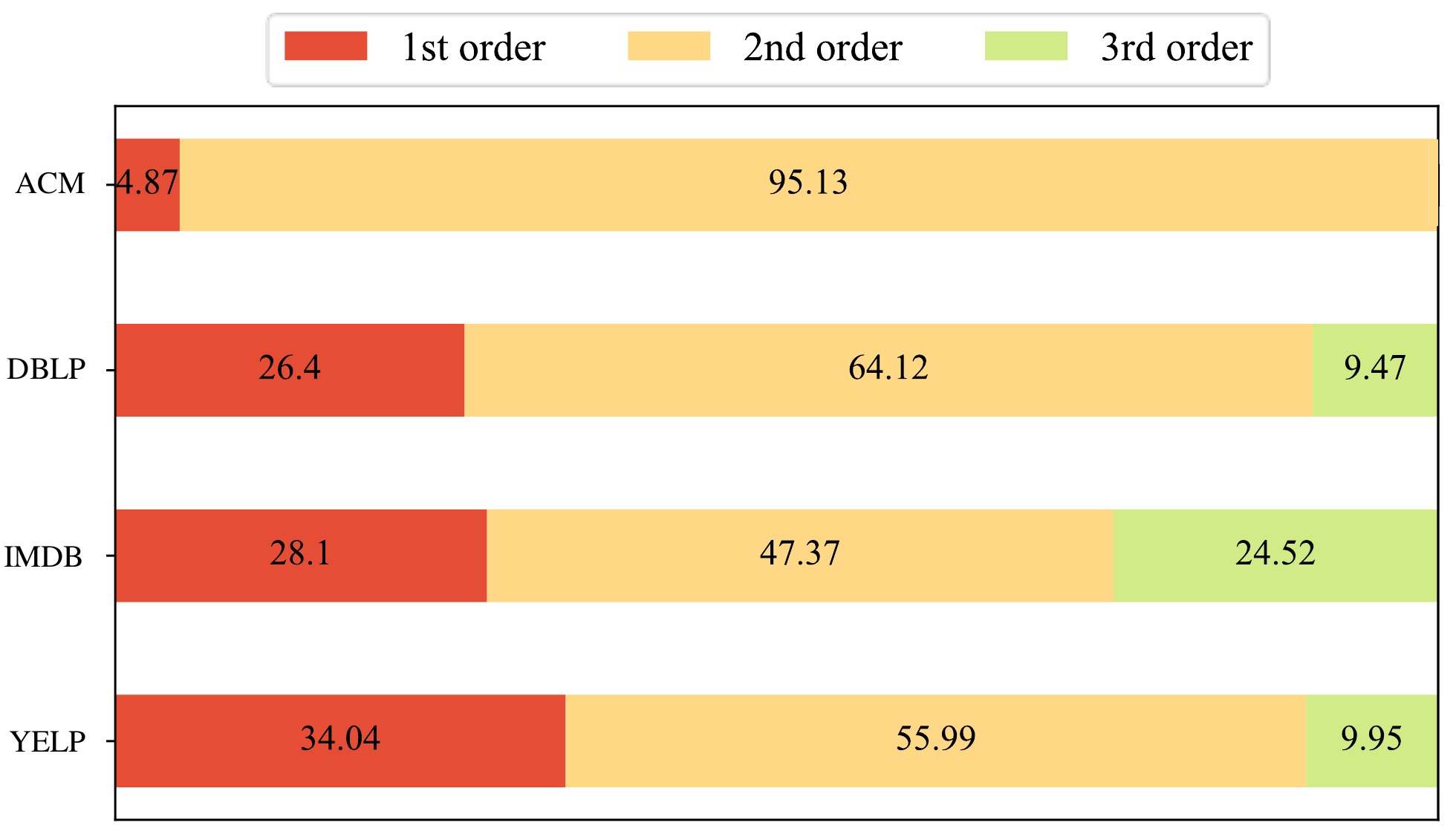}
  \caption{Weights (\%) of different lengths of meta-paths in the multi-order adjacency matrix learned in different datasets.
  }
  \label{WeightsMultiOrder}
\end{figure}

\textbf{Impact of Multi-order Paths}.
Furthermore, we examine the adaptive weights of distinct orders of meta-paths in Figure \ref{WeightsMultiOrder}.
In this paper, we set the highest degree of meta-paths as the number of manually defined first-order meta-paths.
It can be observed that high-order meta-paths play critical roles in the multi-order GCN.
On ACM dataset, the first-order adjacency matrices only take a small percentage of the multi-order meta-paths.
These experimental results indicate that multi-order meta-paths are helpful for the model to capture the information from high-order neighbors, which promote high-quality node embedding learning.

\begin{figure}[!tbp]
  \centering
  \includegraphics[width=0.8\textwidth]{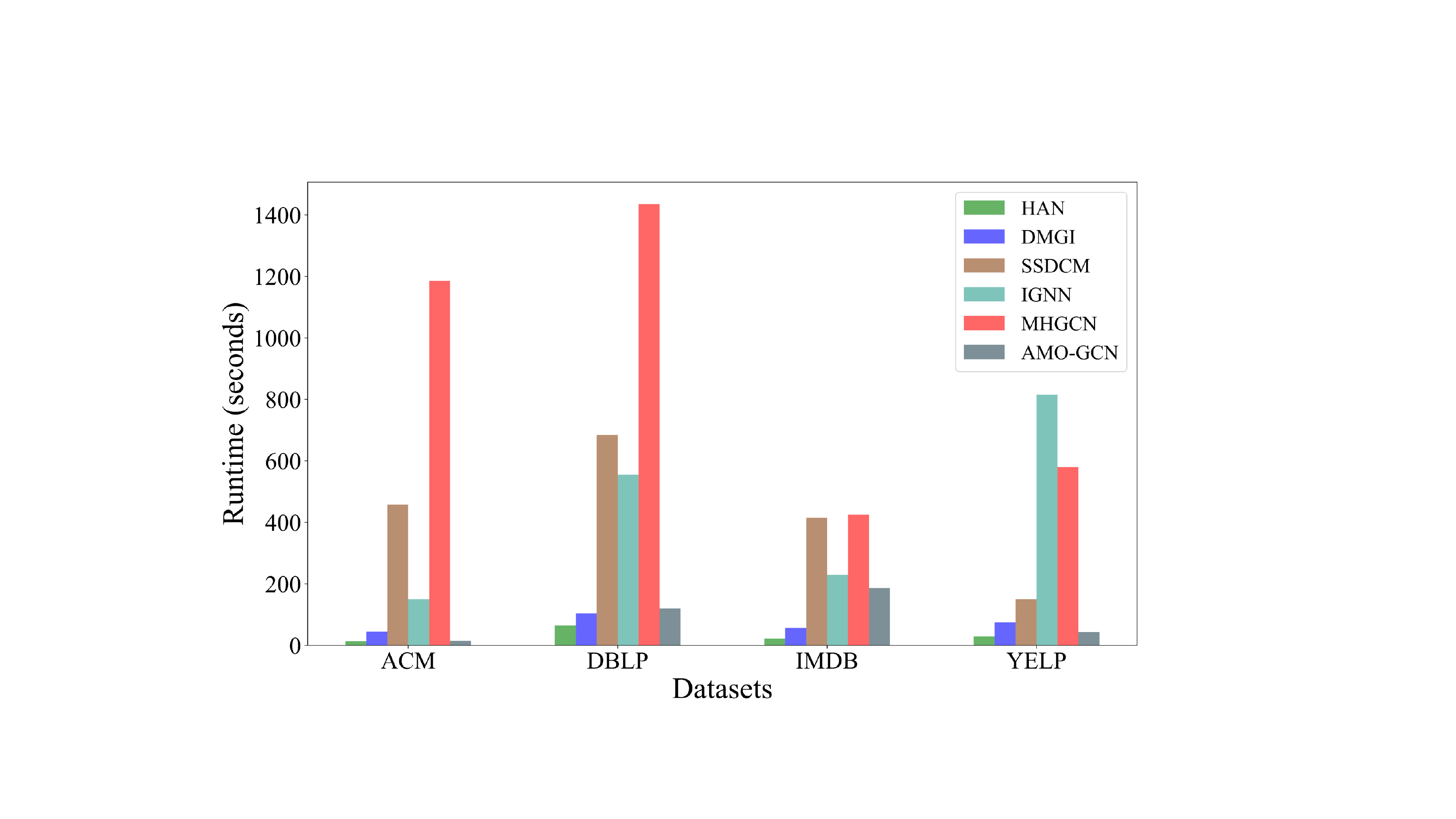}
  \caption{Running time (seconds) of compared HGNNs with 500 training iterations.
  }
  \label{Runtime}
\end{figure}

\textbf{Training Time}.
We compare the training time of HGNNs addressing heterogeneous graphs in Figure \ref{Runtime}.
In general, the runtime of AMOGCN is acceptable compared with HAN and DMGI, and it is more efficient compared with SSDCM and MHGCN.
AMOGCN gains more robust node embeddings with faster speed compared with MHGCN which also adopts learnable high-order meta-paths.
A primary reason for the speed boosting is that we adopt a one-layer graph convolution, rather than using a multi-layer GCN to capture the aggregation of varying lengths of meta-paths as MHGCN.

\begin{figure}[!tbp]
  \centering
  \includegraphics[width=0.8\textwidth]{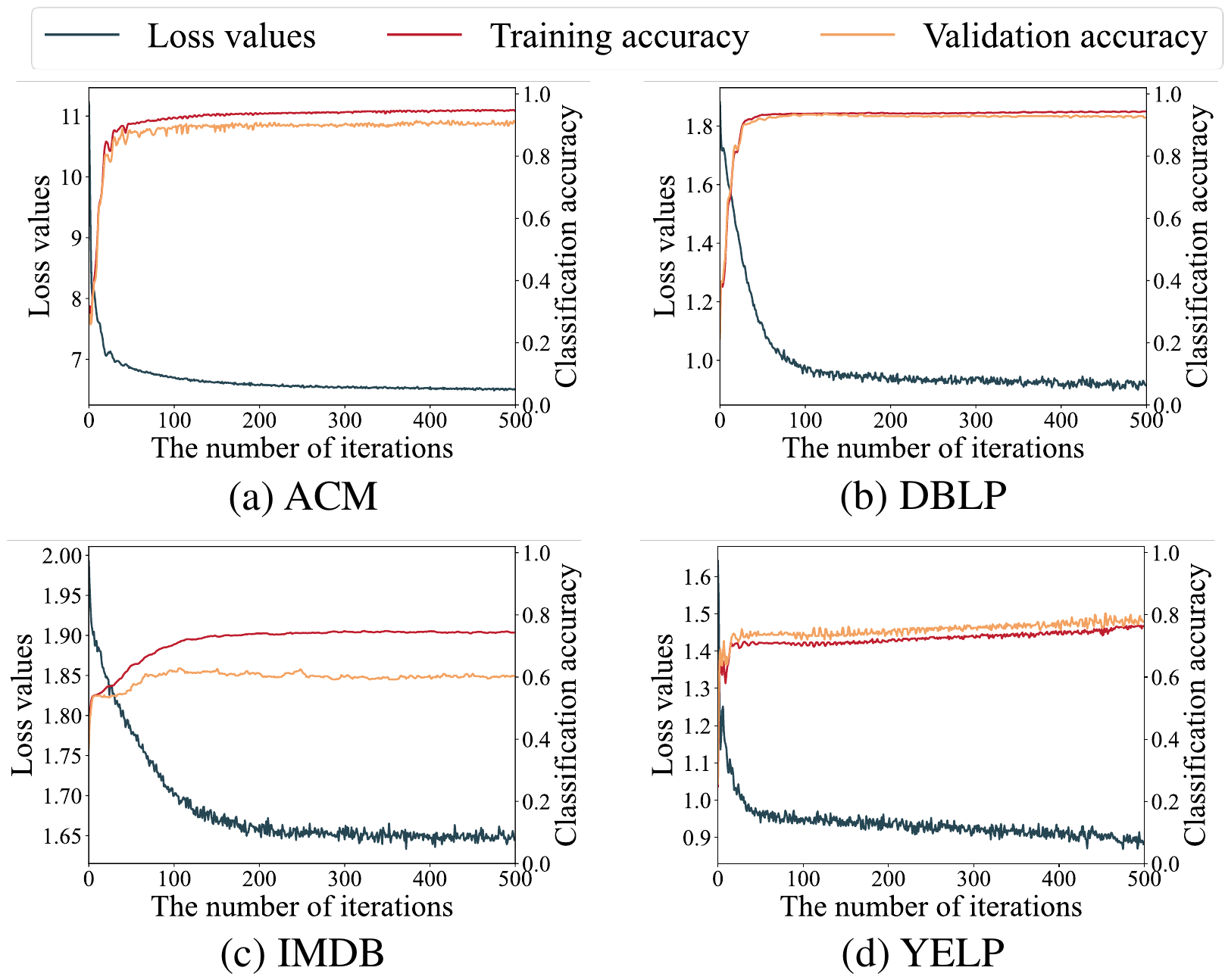}
  \caption{Convergence and training/validation Macro-F1 curves of AMOGCN.
  }
  \label{Convergence}
\end{figure}

\textbf{Convergence Curves}.
Eventually, we record the curves of loss values and accuracy on the training set and the validation set during network learning, as exhibited in Figure \ref{Convergence}. 
It can be observed that loss values decrease rapidly during training and eventually
converge on all datasets.
The accuracy of training and validation sets also grows up as the loss dwindles.
In this paper, we select the optimal model with the highest validation accuracy to get classification predictions on test sets.

\section{Conclusion}\label{conclusion}
In this paper, we proposed an Attributed Multi-Order Graph Convolutional Network (AMOGCN) to cope with heterogeneous networks,
which explored the fusion of different lengths of meta-paths via learning an adaptive multi-order adjacency matrix.
For the purpose of exploring multi-length meta-paths, we first designed the formulation of the adaptive high-order adjacency matrix, which corresponded to a long-length meta-path involving various first-order meta-paths.
AMOGCN further introduced semantic information that considered node homophily as a new type of node relations, which was regarded as supervision signals to instruct the automatic aggregation of multi-order adjacency matrices.
Substantial experimental results pointed out that the proposed model succeeded in learning multi-length meta-paths with a multi-order adjacency matrix, and gained promising performance improvement on heterogeneous graph datasets compared with state-of-the-art competitors.
In our future work, we will devote ourselves to research on joint HGNNs with learnable meta-paths, such as adaptive meta-path refining with more complex heterogeneous graphs.

\section*{Acknowledgments}
This work was partially supported by the National Natural Science Foundation of China (Nos. U21A20472 and 61672159).

\bibliographystyle{elsarticle-num}
\bibliography{MachineLearning}
\end{document}